\documentclass{article}

\usepackage{arxiv}

\usepackage[utf8]{inputenc} % allow utf-8 input
\usepackage[T1]{fontenc}    % use 8-bit T1 fonts
\usepackage{hyperref}       % hyperlinks
\usepackage{url}            % simple URL typesetting
\usepackage{booktabs}       % professional-quality tables
\usepackage{amsfonts}       % blackboard math symbols
\usepackage{nicefrac}       % compact symbols for 1/2, etc.
\usepackage{microtype}      % microtypography
\usepackage{lipsum}		% Can be removed after putting your text content
\usepackage{graphicx}
\usepackage{natbib}
\usepackage{doi}
\usepackage{framed,multirow}
\usepackage{amsmath}
\usepackage[table]{xcolor}

\title{Benchmarking Graph Neural Networks \\ for FMRI analysis}%

\author{Ahmed El Gazzar \\ Amsterdam University Medical Center \\a.g.elgazzar@amsterdamumc.nl \And Rajat Mani Thomas \\ Amsterdam University Medical Center \\rajatthomas@amsterdamumc.nl \And Guido Van Wingen \\ Amsterdam University Medical Center \\guidovanwingen@amsterdamumc.nl}
\begin{document}
\maketitle

\begin{abstract}
Graph Neural Networks (GNNs) have emerged as a powerful tool to learn from graph-structured data. A paramount example of such data is the brain, which operates as a network, from the micro-scale of neurons, to the macro-scale of regions. This organization deemed GNNs a natural tool of choice to model brain activity, and have consequently attracted a lot of attention in the neuroimaging community. Yet, the advantage of adopting these models over conventional methods has not yet been assessed in a systematic way to gauge if GNNs are capable of leveraging the underlying structure of the data to improve learning. In this work, we study and evaluate the performance of five popular GNN architectures in diagnosing major depression disorder and autism spectrum disorder in two multi-site clinical datasets, and sex classification on the UKBioBank, from functional brain scans under a general uniform framework. Our results show that GNNs fail to outperform kernel-based and structure-agnostic deep learning models, in which 1D CNNs outperform the other methods in all scenarios. We highlight that creating optimal graph structures for functional brain data is a major bottleneck hindering the performance of GNNs, where existing works use arbitrary measures to define the edges resulting in noisy graphs. We therefore propose to integrate graph diffusion into existing architectures and show that it can alleviate this problem and improve their performance. Our results call for increased moderation and rigorous validation when evaluating graph methods and advocate for more data-centeric approaches in developing GNNs for functional neuroimaging applications.\footnote{Our code is publicly available at https://github.com/elgazzarr/fMRI-GNNs}
\end{abstract}

% keywords can be removed

\section{Introduction}
Mapping the underlying functional brain activity to observed phenotypes and neurological disorders has been a major goal of neuroimaging research. The emergence of non-invasive imaging techniques such as functional magnetic resonance imaging (fMRI) that estimates the neural activity of the brain by measuring the blood-oxygen level dependence (BOLD) has been paramount in advancing our understanding of the functional organization of the brain\citep{HuettelSongMcCarthy2004}. Yet, modeling the fMRI signal is a challenging feat due to the high dimensionality of the signal coupled with the low sample size, especially in clinical datasets, and the low signal-to-noise ratio of the scans. To build learning models from such data, researchers must opt for tools that incorporate priors from the data structure to guide modeling. This is also called finding the correct inductive bias. Given our knowledge that the brain functions a network \cite{SpornsTononiKotter}, graph-based methods have been instrumental in modeling the brain with the field of network neuroscience utilizing graph theory approaches to find interesting insights about the brain function and dysfunction \citep{Sporns2018}. With the rise of deep learning, GNNs have been developed to extend the merits of neural networks to graph structured data \citep{BronsteinBrunaLeCunSzlamVandergheynst2017}. Naturally, the neuroimaging community has picked up GNNs and applied it to wide range of tasks such as  phenotype prediction \citep{KtenaParisotFerranteRajchlLeeGlocker, ParisotKtenaFerranteLeeGuerreroGlocker}, disease diagnosis \citep{GalloElGazzarZhutovskyThomasJavaheripourMeng, LiDvornekZhuangVentolaDuncan2020} and task inference \citep{LiDvornekZhouZhuangVentolaDuncan2019}. This adoption is still in its infancy, where most of the work is focused on adapting a specific architecture proposed on other domains for a specific neuroimaging application. It therefore remains unclear which graph architecture should be selected. Should we use graph convolution \citep{KipfWelling2016} as \citep{GalloElGazzarZhutovskyThomasJavaheripourMeng} or Graph isomorphic networks \citep{XuHuLeskovecJegelka2018} as in \citep{kim2020understanding}. Should we use graph attention \citep{Velic} as in \citep{HuCaoLiDongLi2021}. Should the input graph be static or dynamic \citep{ElGazzarThomasvanWingen2021}? How to create the edges? What is the sample size suitable for training these models? And most importantly, are these techniques capable of exploiting the underlying structure of the data to improve learning?\\
Despite the theoretical appeal of the application of GNNs to learn from functional brain activity, we find that similar to others domains, the experimental settings for GNNs in neuroimaging applications are in many cases ambiguous or not reproducible \citep{ErricaPoddaBacciuMicheli2019}. Some of the most common reproducibility problems concern hyperparameters selection, where hyperparameters optimization is conducted for the proposed model and not the baselines, or they are optimized based on their performance on the final test set. Another main issue is the incorrect use of data splits for model selection versus model assessment. These limitations could lead to over-optimistic and biased estimates of the true performance of the model, making it difficult to assess the advantage of adopting GNNs for this application or to identify the key components in developing more powerful architectures.\\
With this premise, our primary aim is to provide the neuroimaging community with a fair performance comparison among some of the popular GNN approaches recently adopted for classification of phenotypes or psychiatric disorders from resting-state fMRI scans. We conducted a large number of experiments within a standardized and rigorous model selection and assessment framework for three different large-scale dataset encompassing three different classification tasks. Namely, the \textbf{Rest-meta-mdd} \citep{YanChenLiCastellanosBaiBo} dataset for the classification of major depression disorder, the Autism Brain Imaging Data Exchange (\textbf{ABIDE I+II}) \citep{MartinoYanLiDenioCastellanosAlaerts} dataset for the classification of autism spectrum disorder and the \textbf{UkbioBank} \citep{SudlowGallacherAllenBeralBurtonDanesh} dataset for sex classification.
Secondly, we aim to identify if and to what extent current approaches can effectively exploit graph structures. To this end, we implement three structure-agnostic baselines, whose purpose is to disentangle the advantage of graph structure information from the nodes features. To ensure a fair comparison against GNN methods, we conduct our baseline experiments under a similar framework for hyperparameter optimization and model selection and assessment. 
Our third aim is to study the effect of scaling the sample size on the GNNs and the baselines. We evaluate the performance of the models when trained on different sample sizes on the UkbioBank for sex classification and tested on an independent subset.
Finally and most importantly moving forward, our objective is to apply insights from our implementations and empirical results to highlight strengths and shortcomings of current methods and propose a set of recommendations to help utilize the potential of this powerful set of tools. 

\section{Background}
Graph neural networks are a class of neural networks that extend deep learning to graph structured data, and have been utilized in several domains for tasks such as node classification, edge prediction and graph classification \citep{BronsteinBrunaLeCunSzlamVandergheynst2017, zhou2020graph, wu2019graph}. While graph neural networks have been applied for node classification tasks in functional neuroimaging where the graph represented a population graph \citep{ParisotKtenaFerranteLeeGuerreroGlocker}, a more befitting application is graph classification, where a brain is represented as a mathematical graph $G  =  (V, E, X)$ whose nodes $v_i  \in V$ represent brain regions and edges $E$  constitute their connections. Input graphs for GNN models are often attributed graphs with the attributes representing the node features $x_i  \in X$. A graph can be stationary or dynamic depending on the evolution of the features and/or connections across time. This work addresses the task of graph classification from both static and dynamic functional graphs and using five popular GNN architectures. Our criteria for selection are i) popularity i.e. the architecture has been extensively validated in multiple domains for applications and is commonly used as a benchmark in graph neural networks surveys; ii) strong architectural differences; iii) original code publicly available; iv) previous adaptation on fMRI data. Specifically, we adopt Graph Convolutional Networks (GCNs), Graph Attention Networks (GATs), Graph Isomorphic Networks (GINs) as architectures to learn from static graphs. While for dynamic graphs, we adopt Spatio-Temporal Graph Convolutional Networks (ST-GCNs) and Adaptive Spatio-Temporal Graph Convolutional Networks (AST-GCNs). Several architectural choices for each model are represented as parameters in a search space and are selected using an exhaustive hyperparameter search. Thus each of the models can be viewed as a general class that encompasses several variants of previously proposed GNN models.
We briefly describe the main idea behind each class and some of their previous applications for learning from fMRI data. We discuss the specific details of each method under a uniform framework in Section \ref{gnns}. 

\paragraph{Graph convolutional networks}
A GCN \citep{KipfWelling2016} can be viewed as the counterpart of a convolutional neural network (CNN) model for graph-structured data that uses a graph spectral approach to perform convolution. A graph convolutional layer extracts features through a first-order spectral approximation of the graph by restricting the filters (limiting the order of the Chebyshev polynomial) to operate in the neighborhood at one step away from each node. GCNs are the most popular method applied for graph classification from fMRI data and have been utilized for the diagnosis of psychiatric disorders such as ASD \citep{arya2020fusing, LiDvornekZhuangVentolaDuncan2020, KtenaParisotFerranteRajchlLeeGlocker}, MDD \citep{GalloElGazzarZhutovskyThomasJavaheripourMeng, qin2022using} and Schizophrenia \citep{oh2022diagnosis, lei2022graph}.

\paragraph{Graph attention networks}
GAT \citep{Velic} is a graph neural network architecture that uses the attention mechanism to learn weights between connected nodes. In contrast to GCN, which uses predetermined weights for the neighbors of a node corresponding to the normalization coefficients, GAT modifies the aggregation process of GCN by learning the strength of the connection between neighboring nodes through the attention mechanism on the features of the connected nodes. \citep{LiDvornekZhouZhuangVentolaDuncan2019, LiDvornekZhuangVentolaDuncan2020} have adopted GATs for the diagnosis of ASD.

\paragraph{Graph isomorphic network}
GINs were proposed by \citep{XuHuLeskovecJegelka2018} as a special case of spatial GNN suitable for graph classification tasks. The network uses the sum operation to aggregate features from the neighboring nodes and employs an extra fully connected layer in a non-linear multi-layer perceptron (MLP) to transform the output at each layer. The authors argue the GINs can learn an invective function which deems the model to be possibly as powerful as the Weisfeiler-Lehman test \citep{WeisfeilerLeman1968} for graph classification tasks. \citep{kim2020understanding}) have applied GINs for sex classification from rs-fMRI data.

\paragraph{Spatio-temporal graph convolution network}
ST-GCNs \citep{YuYinZhu2017} has been first proposed to model dynamic data for the application of traffic forecasting. They were developed to extract features from graphs with dynamic attributes that evolve with time and have been popular in several applications such as weather forecasting and skeleton-based activity recognition. A typical ST-GCN block consists of a temporal layer (e.g. 1D Conv layer, LSTM, etc.) to extract temporal features followed by a graph convolution layer for spatial feature extraction. Stacking ST-GCN blocks enables spatio-temporal feature extraction.\citep{GadgilZhaoPfefferbaumSullivanAdeliPohl2020} and \citep{AzevedoPassamontiLioToschi2020} have applied ST-GCNs on fMRI data the tasks of sex and age classification, and \citep{kong2021spatio} for the task of MDD diagnosis. 

\paragraph{Adaptive spatio-temporal graph convolution network}
AST-CGNs \citep{BaiYaoLiWangWang2020} fundamentally share the same concept as ST-GCNs in feature extraction however they differ in the input graph structure. While ST-GCN relies on a predefined adjacency matrix, AST-GCN starts with a randomly initialized graph structure and adaptively learns the optimal graph structure through gradient descent in an end-to-end fashion. While computationally more expensive, this setup enables learning task-specific graph structure in domains where there is no known graph structure or the predefined graph structure is not sufficient. \citep{ElGazzarThomasvanWingen2021} have applied AST-GCN for the tasks of sex and age classification.

\section{Methods}

\subsection{Graph creation}\label{graphc}

This work studies both static and dynamic representation of graphs. Static graphs assume no temporal evolution of the graph structure and nodes features. They are typically created solely using static measures of similarity (e.g. correlations) for both the edges and features. Currently, the most popular approach for static graph creation is using the Pearson correlation of the ROIs time-courses (commonly refereed to as functional connectivity) to define the nodes features as the  connectivity profile of each ROI (the respective row of the correlation matrix) and the graph edges are represented as a sparse version of the correlation matrix. Note that while some works use other approaches such as clustering, structural connectivity and anatomical location for representing graph edges, the use of functional connectivity (FC) to define graph structure remains the most popular approach in the large majority of GNN brain studies.
In this work we follow the same popular strategy to create static graphs and use GCN, GAT, and GIN models for learning from static graphs while treating the sparsfication threshold as a tunable hyperparamter. The threshold is set based on a proportional thresholding (i.e. selecting strongest PT \% connections) rather than absolute threshold as it has shown in graph to have improved test-retest reliability and is less sensitive to variation in graph density \citep{braun2012test, van2010comparing}.
Static graphs are relatively easier to compute, have more training architectures to select from, have a significantly lower number of dimensions and a more robust to physiological noise and inter-scanner variability. However, they fail to represent the dynamic nature of brain activity which is hypothesized to contain the biomarkers for mental disorders \citep{HutchisonWomelsdorfAllenBandettiniCalhounCorbetta2020}. 

\begin{figure*}[!t]
\centering
\includegraphics[width=\linewidth]{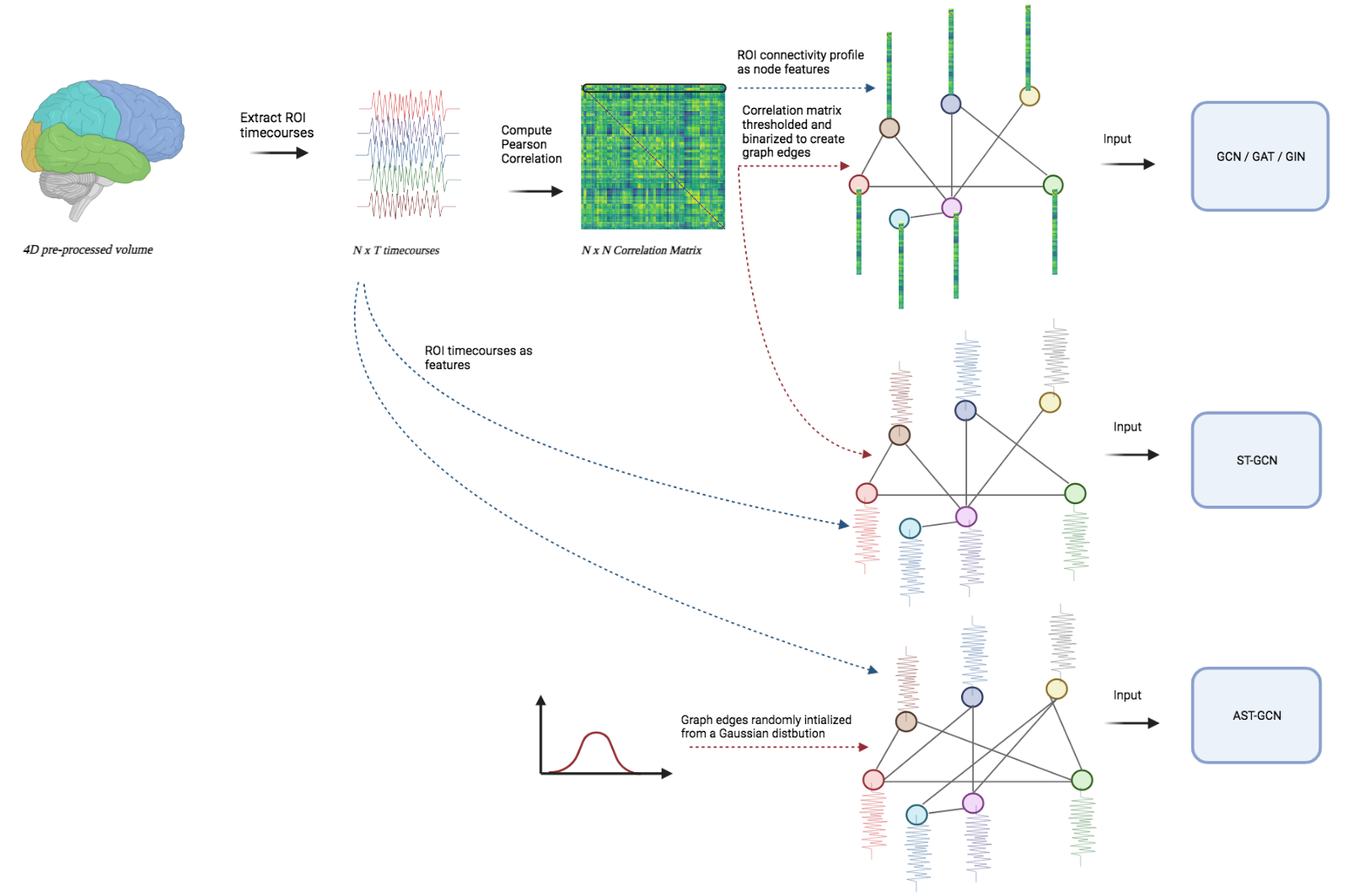}
\caption{\label{fig1} Graph creation strategy for the 5 models employed in this study. Input graphs to GCNs, GATs and GINs solely rely on static functional connectivity to create the nodes features and graph edges. Input graphs to ST-GCNs and AST-GCNs utilize ROIs timecourses to define the nodes features. ST-GCNs graph use thresholded FC to define the graph structure, where AST-GCNs start with a randomly initialized graph and learn the edge weights via gradient descent.}
\end{figure*}

Dynamic graphs enable modelling such dynamic signal by representing the nodes features as ROIs time-courses. Similar to static graphs, dynamic GNNs employ thresholded FC to define the graph adjacency matrices. Another strategy adopted by \citep{ElGazzarThomasvanWingen2021}, is to randomly initialize the adjacency matrix and adaptively learn its weights in an end-to-end manner simultaneous with the GNN weights via gradient descent. We adopt the former strategy for our ST-GCN model and the latter for the AST-GCN model. Figure \ref{fig1} summarizes the graph creation strategy employed for all 5 models in this study.

\subsection{Graph Neural Networks}\label{gnns}

For graph classification, the goal of a GNN is to learn a function $f(.)$ which  maps a graph to a feature vector that summarizes the graph and contains information necessary for classification. The forward function thus can be viewed as : 
\begin{equation}
\centering
y = g(f(G))
\end{equation}

where $y \in R$ is the label of the graph and $g(.)$ is a function parameterized by a neural network to map the output feature vector of the GNN to the label. In this work, we choose $g(.)$  as a single-layer fully connected neural network for all the different GNN architectures. Thus, the empirical results would only reflect how powerful are different parameterizations of $f(.)$ in representing the graph signal and structure. In this section, we explain how each GNN architecture models this function. 

\subsubsection{GCNs}

A GCN learns graph representations via an approximation of spectral convolutions on the graphs. Spectral convolutions on graphs is defined  as the multiplication of the the graph signal $x$ with a filter $s_{\theta}$ in the Fourier domain. 

\begin{equation}
    s_{\theta} \ast x = Us_{\theta}U^{T}x 
\end{equation}

where $U$ is the matrix of eigenvectors of the normalized graph Laplacian $L= I - D^{-1/2}A D^{-1/2}=U\Lambda U^{T}$   with a diagonal matrix of its eigenvalues $\Lambda$ and $U^{T}x$  being the graph Fourier transform of $x$. Evaluating this spectral convolution in its current form is computationally expensive because of the multiplication by $U$ and the eigendecomposition of  $L$.  \citep{hammond2011wavelets} suggested that $s_{\theta}$  can be well-approximated by a truncated expansion in terms of Chebyshev polynomials up to K-th order. Follow up work by \citep{KipfWelling2016} proposed to restrict the polynomials to their first order for each graph convolutional layer and stacking $K$ graph convolutional layer to recover a rich class of convolutional filters.  The layer-wise propagation rule of the GCN can thus be defined in the spatial domain as : 

\begin{equation}
    H^{k+1} = \sigma(\hat{D}^{-1/2}\hat{A}\hat{D}^{-1/2}H^{k}W^{k})
\end{equation}

where $H^k$  is the graph signal representation at layer $k$ ; $H(0)=X$, $\hat{A}$ is the adjacency matrix with self-connections, $\hat{D}$ is the degree matrix, $W(k)$  is a trainable weight matrix and $\sigma$ denotes a non-linear activation function. \citep{gilmer2017neural} later formalized this definition as message passing and update operations on the nodes. The node-wise update rule for a GCN can thus be defined as 

\begin{equation}\label{gcn-eqn}
    h^{k+1}_{i} = \phi(h^{k}_{i}, \sum_{j \in \mathcal{N}_i}c_{ij} \psi(h^{k}_j))
\end{equation}

where $\psi$ and $\phi$ are the message and update functions respectively, $c_{ij}$ is the edge weight between nodes $i$ and $j$ (set to 1 in unweighted graphs) and $\mathcal{N}_i$ is the neighborhood of the node $i$. In this work, we set $K=2$ as we observed no additional gain from convolving higher-order neighborhoods of the graph nodes. $\psi$ is a linear fully connected layer $W$ and $\phi$ is a is a single-layer feed forward network with a hidden dimension $d_h$ set as a hyperparameter with a non-linear ReLU(·) activation function.
 
\subsubsection{GATs}

The main difference between a GCN and GAT layer lies in the permutation invariant update function $\phi(h_i, h_{\mathcal{N}_i})$ applied to each node i to combine the node features hi with its aggregated neighborhood features $h_{\mathcal{N}_i}$  (transformed by some function $\psi$) and update its representation. In GCNs, importance of node j to node i’s representation is defined using $c_{ij}$, the entry $\hat{A}_{ij} $on the normalized adjacency matrix. For GATs, the importance is learned implicitly using a self-attention mechanism. Formally, the update rule for each node in a GAT layer can be defined as:
\begin{gather}
    h^{k+1}_{i} = \phi(h^{k}_{i}, \sum_{j \in \mathcal{N}_i}a(h^{k}_i, h^{k}_j)\psi(h^{k}_j)) \\
    a(h^{k}_i, h^{k}_j) = \textit{Softmax}(\sigma(m^{T} [ Wh^{k}_i\vert\vert Wh^{k}_j]))
\end{gather}

$\vert\vert$ denotes a concatenation operation, $\sigma$ is a non-linear activation function and $m$ is a trainable single-layer feed-forward neural network. This attention mechanism can be extended to a multi-head attention via learning M different copies of $a(h^k_i , h^k_j)$ and calculating their average. In this work, we treat $M$ as a hyperparameter with an option of setting $M=1$ given that the sample size is limited and increasing the model trainable parameters by a large factor can lead to over-fitting. We follow the author’s recommendation of choosing $\sigma$ as LeakyReLU( ) and parameterizing the linear transformation of the feature vectors via a linear layer. For the rest of the parameters of the GAT model (Number of layers, size of the hidden dimension and graph readout layer), we follow the same setting as described in Section 4.1.1

\subsubsection{GINs}

The Graph Isomorphism Network (GIN) is a variant of the GNN suitable for graph classification tasks, which is known to be as powerful as the WL-test under certain assumptions of injectivity. Unlike GCNs and GATs which can use any permutation invariant function to aggregate neighborhood features, GINs typically only use the sum operation and do not include a transformation function $\psi$ before combining the neighborhood features with the original node features. Rather, a multi-layer perceptron (MLP) is applied directly on the combined features. The update rule for each node in a GIN layer be defined as:
\begin{equation}
    h^k_{i+1} = \textit{MLP}( (1+ \epsilon ) h^k_i +  \sum_{j \in \mathcal{N}_i}h^k_j)
\end{equation}

where $\epsilon$ is a learnable parameter initialized to zero. The purpose of this parameter is to learn the amount of original information in the node that it should retain in the next representation. We use a two layer $MLP$ with ReLU() as non-linearity and as recommended by the paper.  For the READOUT layer, we experiment with the three permutation invariant functions as in Section 4.1.1.

\subsubsection{ST-GCNs}

ST-GCNs model graphs with dynamic features by utilizing spatio-temporal blocks that alternate between sequence-to-sequence layers for learning temporal features, and graph convolutions for spatial feature extraction. While there has been several variants of ST-GCNs that differ either in the number or order of layers in the ST-GCN block, or the the temporal learning layer (e.g., gated temporal convolutions, dilated convolutions, LSTMs), we follow the work by \citep{wu2019graph} to implement our variant of ST-GCN for fMRI as we have observed it provided the most stable results across tasks. Namely, we utilize a cascade of $M$ St-GCN blocks for spatio-temporal feature extraction, where each ST-GCN block consists of a one layer of gated temporal convolution (GATED-TCN) for temporal feature extraction followed by a spatial graph convolution as described in Section for spatial feature extraction. 
For a node input representation $x_{i} \in R^{T \times C}$ where $C$ is the number of channels, a Gated-TCN extracts temporal feature representation $h_{i}$ as follows:

\begin{equation}
    h_{i} = \textit{tanh}(W_a \ast x_i + b) \odot \textit{Sigmoid}(W_b \ast x_i + c) 
\end{equation}

where $W_a$ and $W_b$ are the weights of two parallel 1D convolutions for controlling the flow of information into the next layer, $b$ and $c$ are the biases and $\odot$ represent the Hardmard product operation. The extracted feature representations are then treated as the nodes input representation in a GCN and is processed similar to Equation \ref{gcn-eqn}. Another key difference between the ST-GCN employed this work and prior implementations of ST-GCNs for modelling fMRI data is that we use the entire timecourses for the duration of the scan as the node features as opposed to previous works by \citep{GadgilZhaoPfefferbaumSullivanAdeliPohl2020} \citep{AzevedoPassamontiLioToschi2020} who use crops of $t=30-60s$. We found that cropping the timecourses for training and aggregating the predictions during inference drastically deteriorates the performance for the clinical tasks.

\subsubsection{AST-GCNs}
In multiple spatio-temporal datasets, we know that the underlying interaction between the temporal features can be modelled as a graph, but we don't have access to the ground truth structure of the graph. Adaptive learning of graph structure is a growing area of research in the field of GNNs and several variants have been proposed by the research community in recent years \citep{BaiYaoLiWangWang2020}, \citep{wu2019graph}, \citep{wu2020connecting}. AST-GCNs can be generalized as ST-GCNs equipped with a graph structure learning module. In this work, we follow the implementation of \citep{BaiYaoLiWangWang2020} which defines the normalized adjacency matrix as 

\begin{equation}
    \hat{A} = I_N + \textit{Softmax}(\textit{ReLU}(E \cdot E^{T}))
\end{equation}

where $E \in R^{N \times d_e}$ is a learnable node embedding dictionary initialized randomly with dimension $d_e$ as a hyperparameter. Multiplying the dictionary by its transpose results in a dense symmetric $N \times N$ matrix, which is then passed through a $ReLU$ activation function to eliminate weak connections and then row-normalized using the \textit{Softmax} operation. We also experimented with the \textit{Sparsemax} operation \citep{martins2016softmax} and found that it improved stability during learning. The rest of our AST-GCN implementation is kept similar to our ST-GCN to disentangle the effect of adaptive learning of the graph structure.

\subsection{Graph Readout}
Following feature extraction using any of the proposed GNN architectures, 
the readout phase computes a feature vector for the whole graph using some permutation-invariant readout function to be passed as an input to the final classification layer. In this work we experiment with 3 different permutation-invariant functions to summarize the graph features after the last graph layer. Namely \textit{Mean()}, \textit{Mean()$\vert \vert$Max()} and \textit{Sum()}. The selection of which function to use for each architecture/dataset combination is set as a hyperparameter. We empirically observe that \textit{Mean()} and \textit{Mean()$\vert \vert$Max()} provide the most stable results to all GNN architectures with exception of the GIN architecture where the \textit{Sum()} operation marginally improve the performance of the model.

\section{Data \& Experimental Setup}

\subsection{Datasets}

Throughout this study we evaluate the performance of the models on three different datasets, each with a different classification target and a different prepossessing pipeline. This setup enables us to make a generalized conclusion from the empirical results that are robust to variation in demographics, classification target, pre-processing pipeline and data acquisition protocols.

\textbf{ABIDE I+II} contains a collection of rs-fMRI brain images aggregated across 29 institutions. It includes data from participants with a clinical diagnosis of autism spectrum disorder and typically developing participants (TD). In this study, we select a subset of the dataset with N = 1207 (558 TD/649 ASD) using the largest nine sites in terms of number of participants (N$\geq$50) to conduct our analysis. The 4D volumes were pre-processed according to the CPAC pipeline \citep{CraddockSikkaCheungKhanujaGhoshYan} with the same parameters as in \citep{CraddockBenhajaliChuChouinardEvansJakab}. Further, the craddock-200 \citep{CraddockJamesIiiHuMayberg2012} atlas was used to parcellate the volume into 200 regions of interest (ROIs) and the timecourses of each region were calculated as the mean signal between all the voxels of the ROI.

\textbf{Rest-meta-MDD} is currently the largest open-source rs-fMRI database for studying major depression disorder including clinically diagnosed patients and healthy controls from 25 cohorts in China. In this study we select a subset of the dataset with N = 1453( 628 HC/825 MDD) using the largest 9 sites (N $\geq$ 40). Standard pre-processing of the data was done at each site using the Data Processing Assistant for Resting-State fMRI (DPARSF) \citep{YanZang2010} and 112 timecourses of cortical and sub-cortical regions as defined by the Harvard-Oxford atlas \citep{MakrisGoldsteinKennedyHodgeCavinessFaraone} were extracted.

\textbf{UkBioBank} is a large-scale biomedical database and research resource, containing in-depth genetic and health information  from  half  a  million  UK  participants.  In  this  work  we  use  a  subset  with (N=5500 (2750 M/ 2750 F)) randomly sampled from the first release. The dataset provides the pre-processed 4D volumes in MNI format. These volumes were then parcellated to 116 anatomical regions using the automated anatomical labeling (AAL) atlas \citep{TzourioMazoyer} and the mean timecourse of each region was extracted.

\subsection{Model selection \& assessment}
For a fair evaluation of the methods, we constructed a systematic pipeline to find the optimal performance of each of the methods. This constitutes using a configuration of hyperparameters for each model that are optimal for the given task. A common pitfall in hyperparameter tuning is conducting the search based on the performance of the final test set. This results in an over-optimistic estimate of the model generalization capability. In this work, we conducted hyperparameter search for each GNN model on a held out validation set for each dataset. We conducted an exhaustive hyperparameter search on the key parameters that could influence the convergence of the model and its performance. While some of the parameters are exclusive to some architectures (e.g. the number of attention heads in the GAT), the majority of hyperparameters and their search spaces were shared across methods (See Table S1 for details on the hyperparameters and their search spaces). This enables a relatively fair evaluation of each architecture. After finding the best hyperparameter configuration for each model and the respective tasks, we followed a 5-fold cross validation scheme with model selection using the lowest loss value on an inner validation split (85/15  training/validation split) for the Rest-meta-MDD  and ABIDE I+II  datasets. For the UkBioBank dataset, the models are trained on subsets of the full sample (N = [500, 1000, 2000, 5000]) with a similar inner validation split for model selection and the test results are reported on a fixed independent set of N = 500 to evaluate the effect of training sample size on the performance. (See Supplementary information for details about training procedure)

\subsection{Non-graph baselines}
To fairly judge whether GNNs are fully capable of leveraging the graph structure, their performance need to be evaluated against baselines that are not inductively aware of the structure of the data, and a similar framework should be followed for hyperparameter search and model selection. In this work we employ three different structure-agnostic baselines. Specifically, we adopt a multi layer perceptron (MLP), support vector machines with radial basis function kernel (SVM-rbf) as baselines for learning from static representations. The input to these models are the flattened lower-triangular matrix of the correlation matrix. For a baseline that learns from dynamic representations, we used 1D CNNs similar to \citep{el2019simple}. The inputs to the 1D CNN are the prepossessed ROI time-courses stacked as channels. \textit{(See Supplementary information for details about training the baselines and their hyperparameters)}

\begin{table*}[!t]
\caption{\label{tab1}5-fold cross validation results (balanced accuracy, sensitivity, specificity) for GNNs and baselines on the two clinical tasks of MDD and ASD diagnosis.}
\centering
\begin{tabular}{|c|l|lll|lll|}
\hline
\multicolumn{1}{|l|}{}                  & \multicolumn{1}{c|}{}                                  & \multicolumn{3}{c|}{\textit{\textbf{Rest-meta-MDD}}}                                                  & \multicolumn{3}{c|}{\textit{\textbf{ABIDE I+II}}}                                                     \\ \cline{3-8} 
\multicolumn{1}{|l|}{}                  & \multicolumn{1}{c|}{\multirow{-2}{*}{\textbf{Models}}} & \cellcolor[HTML]{CCCCCC}Acc. \% & \cellcolor[HTML]{CCCCCC}Sens. \% & \cellcolor[HTML]{CCCCCC}Spec. \% & \cellcolor[HTML]{CCCCCC}Acc. \% & \cellcolor[HTML]{CCCCCC}Sens. \% & \cellcolor[HTML]{CCCCCC}Spec. \% \\ \hline
                                        & \cellcolor[HTML]{F3F3F3}GCN                            & 62.6 ± 3                        & 57.6 ± 8                         & 67.8 ± 8                         & 67.6 ± 2                        & 69.5 ± 4                         & 65.7 ± 3                         \\
                                        & \cellcolor[HTML]{F3F3F3}GIN                            & 62.7 ± 3                        & 67.3 ± 5                         & 58.1 ± 3                         & 68.5 ± 1                        & 81.7 ± 3                         & 55.4 ± 4                         \\
\multirow{-3}{*}{\textit{Static GNNs}}  & \cellcolor[HTML]{F3F3F3}GAT                            & 62.9 ± 2                        & 69.5 ± 3                         & 56.4 ± 5                         & 68.2 ± 2                        & 63.5 ± 5                         & 72.9 ± 3                         \\ \hline
                                        & \cellcolor[HTML]{F3F3F3}ST-GCN                         & 58.2 ± 4                        & 48.6 ± 8                         & 67.8 ± 9                         & 65.3 ± 2                        & 67.2 ± 4                         & 63.4 ± 4                         \\
\multirow{-2}{*}{\textit{Dynamic GNNs}} & \cellcolor[HTML]{F3F3F3}AST-GCN                        & 60.7 ± 2                        & 44.6 ± 2                         & 75.8 ± 8                         & 67.8 ± 2                        & 70.8 ± 5                         & 64.9 ± 3                         \\ \hline
                                        & \cellcolor[HTML]{F3F3F3}1D CNN                         & 63.2 ± 3                        & 68.5 ± 5                         & 59.6 ± 6                         & 70.6 ± 2                        & 72.7 ± 3                         & 68.5 ± 3                         \\
                                        & \cellcolor[HTML]{F3F3F3}SVM-rbf                        & 61.9 ± 2                        & 79.0 ± 5                         & 44.9 ± 5                         & 69.5 ± 2                        & 75.9 ± 3                         & 63.0 ± 3                         \\
\multirow{-3}{*}{\textit{Baselines}}    & \cellcolor[HTML]{F3F3F3}MLP                            & 58.4 ± 4                        & 60.1 ± 2                         & 56.8 ± 3                         & 66.6 ± 2                        & 72.1 ± 1                         & 61.2 ± 2                         \\ \hline
\end{tabular}
\end{table*}

\section{Results}

\subsection{Disorder diagnosis}
The results in Table \ref{tab1} show the test performance of the GNNs and baselines on the clinical datasets. On the clinical tasks (N=1207 and N=1453), all the GNNs perform competitively where no one architecture significantly superior to the others. A notable observation is that static GNNs marginally outperformed dynamic GNNs. This contradicts the main hypothesis that dynamic representations are more faithful to natural dynamics of brain activity over summarized static correlation values and thus would result in better results. This is however true for the dynamic baseline, where the 1D CNN outperformed all the methods on both tasks. In general, graph-agnostic baselines performed on par with or marginally outperformed GNN methods despite having no awareness of the structural organization of the data. These empirical results raise even more questions; do GNNs simply fail to leverage the graph structure in learning better representations and hence obtain superior results, or do the heterogeneity of the data and possible noise in the clinical labels create an upper bound on the performance of the models, rendering quantitative evaluation of the merits of GNNs challenging? How does the sample size affect the performance of these models?

\subsection{Scaling performance on the UKBioBank}

\begin{figure*}[!t]
\centering
\includegraphics[scale=.5]{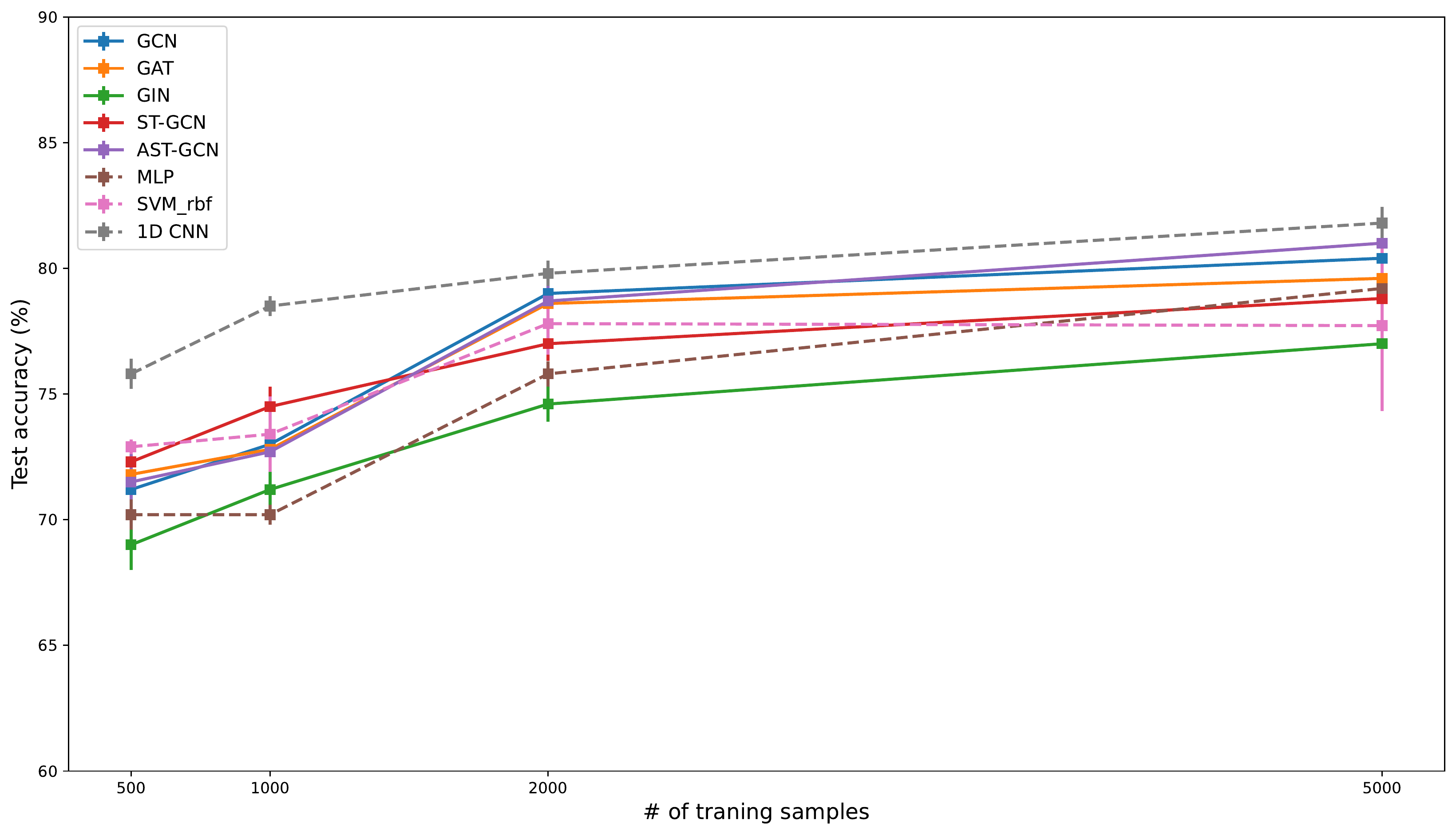}
\caption{\label{fig3}Scaling performance of GNNs and Baselines on the task of sex classification at different available training samples of the UkBioBank dataset.}
\end{figure*}

To shed light on some of these questions, we can look at the scaling performance of these models on the UkBioBank at different training samples on the task of sex classification. Figure \ref{fig3} shows that, similar to the clinical datasets, 1D CNN outperform the rest of the methods at all samples instances. This is especially clear on the smaller sample sizes (N=500 and N=1000). This gap is then reduced at the larger sample sizes, where the dynamic AST-GCN model show improved scaling performance and eventually catch up to the 1DCNN. Again, while these results emphasize the benefits of incorporating the dynamic signal, it questions the sample-efficiency of GNN models in clinical applications, especially since the motivation behind adopting GNNs is that incorporating inductive bias about the structure of the data should alleviate the need for larger sample sizes to learn this solely from data. An advantage we unfortunately do not observe in our empirical results.

A key property of AST-GCN is that it does not require specifying the graph structure in advance, and is jointly learned using gradient descent along with the rest of the model at the cost of extra training parameters. The scaling behaviour of AST-GCN is thus empirically justified and similar scaling performance was previously reported for adaptive spatio-temporal GNNs \citep{ElGazzarThomasvanWingen2021, wu2020connecting}. The performance of AST-GCN however, attest against the correctness of the graph structure of the data used as input to the rest of the GNN models. As we highlighted in Section \ref{graphc}, unlike several graph-structured data, there is no ground truth network structure for the brain and thus defining adjacency matrix of fMRI graphs is rather an arbitrary process with most works opting for FC for this role. In this work, we followed a similar strategy in effort to evaluate the performance of GNN models. The results of our experiments invites us to rethink about this aspect and to find more principled approaches to accommodate this problem. The results of AST-GCN guide toward end-to-end learning of the graph structure as a potential solution to this problem. This however, comes at the expense of large training samples, a luxury not available in most of clinical datasets.

% \usepackage{multirow}

\begin{figure}[!t]
\centering
  \includegraphics[scale=.85]{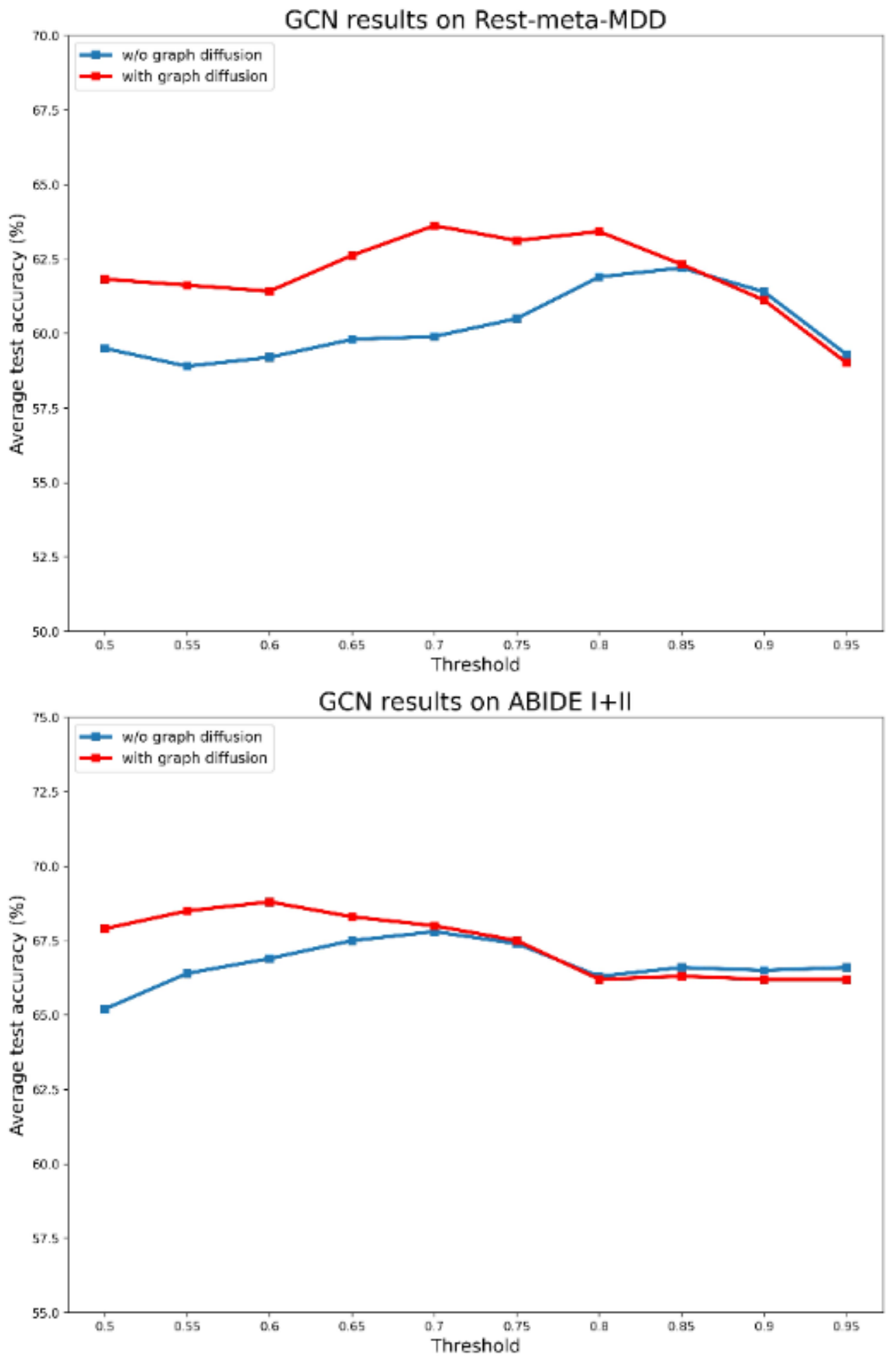}
  \caption{Average test accuracy of the GCN model on the two clinical datasets at different sparsfication thresholds for creating the graph with and without graph diffusion.}\label{difffig}
\end{figure}

\subsection{Graph diffusion improves learning on noisy graphs}

In network neuroscience, data-driven analysis on thresholded connectivity graphs are acknowledged to be potentially problematic, where different thresholding techniques and values and can lead to the inclusion of spurious connections confounding the analysis \citep{van2017proportional, van2010comparing, fornito2012schizophrenia}. Yet we argue that this remains an overlooked problem when using GNNs. During the course of this work, we treated the sparsfication proportional threshold as a hyperparameter that is selected based on the best performance on a validation set. While conducting the experiments, we observed a high sensitivity of the results to the sampled sparsfication threshold. Further, we observed different optimal thresholds for each task and GNN class. In Figure \ref{difffig}, we present the results of the GCN model on the two clinical datasets with all parameters fixed and using sparsfication thresholds in the range of [50, 95] \% with a step size of 5 \%. The empirical results show that selecting an optimal threshold plays an important role in the final evaluation performance and attest against arbitrary threshold selection (e.g. 50 \% as is the current practice) in creating input graphs for GNNs. Further, improved performance at higher thresholds (70-85\%) highlights that proportional thresholding of FC results in noisy graphs. In graph theoretical analysis, proposals to mitigate this problem are to include graph density as a covariate or to adjust the null distribution to account for random edges resulting from thresholding low density graphs \citep{van2017proportional}. An open question is how to deal with such noisy FC graphs while training GNNs on fMRI data.

A possible solution to dealing with noisy graphs is graph diffusion convolution (GDC) \citep{klicpera2019diffusion}. Where instead of aggregating information only from the first-hop neighbors, GDC
models message passing between nodes as a generalized continuous diffusion process defined by the diffusion matrix 
\begin{equation}\label{diff}
    S = \sum_{k=0}^{\infty} \theta_{k}T^{k}
\end{equation} 

with the weighting coefficients $\theta_{k}$, and the generalized transition matrix $T$. In principle, one can choose different methods to define the coefficients and the transition matrix provided that Eq. \ref{diff} converges. Diffusion coefficients can be defined using well-studied functions such as the heat kernel \citep{kondor2002diffusion}, personalized page rank (PPR) \citep{brin1998anatomy} or even learned using gradient descent. Popular examples of defining the transition matrix are the random walk transition matrix $T_{rw} = AD^{-1}$ and the symmetric transition matrix $T_{sym} = D^{1/2}AD^{-1/2}$. Essentially by restricting the sum in Eq. \ref{diff} to a finite number of steps $K$, and choosing suitable diffusion parameters, GDC transforms the original potentially noisy adjacency matrix $A$ into a new graph $S$ where any GNN model can be applied. This transformation has been shown to consistently improve the performance of GNN models on several real world graph datasets. In Fig \ref{difffig}, we show the effect of applying graph diffusion using the heat kernel coefficient and the symmetric transition matrix with $K=2$ on the GCN model on the two clinical datasets at different sparsification thresholds compared to using the original defined graph structure without any transformation (i.e. setting $K=1$, $\theta_{k} =1$ and $T=T_{sym}$). The results show that by simply incorporating this transformation as a prepossessing step, the performance of the model is less sensitive to the sparsfication threshold at lower thresholds and can even improve the empirical results without the need for expensive hyperparameter search for the optimal sparsfication threshold.

\section{Discussion}

The rapid pace of adoption of GNNs in neuroimaging applications urges us to pause and ponder about the rationale and success of this adoption. While we agree that the motivation is well-grounded in the field and in fact necessary for sample-efficient factual learning models, we question whether the current methods are living up to the promises of the motivation. We argue that it is difficult to answer this question without a systematic evaluation of the methods and their current alternatives under a uniform fair setup. In this study, our aim was to build such a framework by benchmarking popular classes of GNN architectures through the applications of brain disorder diagnosis and phenotype prediction. The choice of the application is motivated by the trend and interest of the researchers to use GNNs for such tasks and the open-source nature of the datasets.
In building our framework, we strive for a rigorous fair and setup through i) uniform model selection and assessment strategies across methods, ii) fair allocation of computational resources to all the methods by providing an exhaustive range of parameter search spaces for each architecture/task combination to obtain the best possible performance for each method  iii) uniform training setup (e.g. loss functions, regularization, initialization schemes) for all deep learning methods to disentangle only architectural effects on the performance. 
Our empirical results show that, on average, GNNs fail to outperform current learning alternatives, and that and no one GNN architecture was consistently superior over the others. A simple 2 layer 1D CNN where all the ROIs stacked as channels with no structural awareness of the graph consistently outperformed sophisticated structurally aware GNN methods. The results of AST-GCN -- a dynamic model that learns the graph structure at the cost of more training parameters -- showed an improved performance at the largest sample size over FC-driven graphs. This opens the question whether graph structures defined using thresholded functional connectivity, as is done throughout this work and as is the current practice in the field, is the most optimal strategy for constructing the graphs. We highlighted that thresholded FC is a major issue when computing graph theory measures in network neuroscience, yet, remains an unacknowledged problem in GNNs research. We proposed to integrate graph diffusion as a prepossessing step and show that it can partially alleviate the problem and consistently improve the results. While not the only solution, this calls for a more data-centeric approach when developing learning models from fMRI graphs over a model-centeric approach where we observed that architectural differences did not play an important role in the final outcome in the studied applications. 
On a more general note, the empirical results open the debate of the utility of deep learning in neuroimaging applications \citep{he2020deep, schulz2020different}. On the clinical datasets, the SVM model performed competitively with the DL models without the need for engineering complex neural network architectures. We argue that the competitiveness of DL methods at this scale is a positive rather than a criticism. On the UkBioBank, the dynamic DL models (1D CNN, AST-GCN) do show improved performance over the SVM and other static baselines. This guides towards developing DL models that utilize minimally pre-processed dynamic signal over static feature engineered summaries such as correlation values.  Further, to objectively identify the merits of applying GNNs for single-subject prediction we have to look further than the test metrics and consider the advantage of enabling the application of deep learning methods in this domain. DL is a versatile class of algorithms with applications that extend beyond supervised classification and would potentially be of remarkable value when applied on fMRI data. For example,  graph explainability techniques \citep{ying2019gnn, huang2022graphlime} could be utilized to find potential functional connectivity biomarkers for phenotypes and psychiatric disorders. The application of graph self-supervised learning techniques \cite{fedorov2021self, wang2022contrastive,peng2022gate} and graph normative modelling \citep{gazzar2022improving} on non-clinical datasets would assist in learning representations that could further be transferred to a downstream clinical tasks. Graph augmentation methods \cite{zhao2021data} could boost the statistical power of the clinical datasets and assist in solving the problem of labeled data scarcity and regularize overfitting. These are some of the promising directions that support the utility of GNNs in  neuroimaging. That said, the empirical results presented in this work advocate for more moderation and more rigorous validation when reporting GNN results that do not clearly outperform graph-agnostic baselines. Finally, we hope that framework developed throughout this work could serve as a baseline for future novel GNN architectures or applications on fMRI data and assist in identifying key components that could improve existing methods.

\section*{Acknowledgments}
This work was supported by the Netherlands Organization for Scientific Research (NWO; 628.011.023); Philips Research; ZonMW (Vidi; 016.156.318). The access of the UKbioBank data was granted under the application number 30091.

\bibliographystyle{unsrtnat}
\bibliography{references} 

\end{document}